\documentclass{article} 
\usepackage{iclr2024_conference,times}

\usepackage{amsmath,amsfonts,bm}

\def\eqref#1{equation~\ref{#1}}

\def\1{\bm{1}}

\DeclareMathAlphabet{\mathsfit}{\encodingdefault}{\sfdefault}{m}{sl}
\SetMathAlphabet{\mathsfit}{bold}{\encodingdefault}{\sfdefault}{bx}{n}

\usepackage{hyperref}
\usepackage{url}

\title{The Wisdom of Partisan Crowds:\\ Comparing Collective Intelligence in Humans and LLM-based Agents}

\author{Yun-Shiuan Chuang, Siddharth Suresh$^\dagger$, Nikunj Harlalka$^\dagger$, Agam Goyal\\
  \textbf{Robert Hawkins, Sijia Yang, Dhavan Shah, Junjie Hu, Timothy T. Rogers} \\
  University of Wisconsin-Madison\\
  \texttt{\{yunshiuan.chuang,siddharth.suresh,nharlalka,agoyal25\}@wisc.edu}\\
  \texttt{\{rdhawkins, syang84, dshah, junjie.hu, ttrogers\}@wisc.edu}\\
  \footnotesize$^\dagger$ joint second author
  }

\usepackage{graphicx}
\usepackage{soul}
\usepackage{subcaption}
\usepackage{booktabs}
\usepackage{amsmath}
\usepackage{makecell}
\usepackage{mdframed}
\usepackage{xspace}
\usepackage{wrapfig}

\newcommand{\pb}{\overline{\beta_{\text{PB}}}\xspace}
\newcommand{\woc}{\overline{\Delta \varepsilon}\xspace}

\newcommand{\pbq}{\beta_{\text{PB}_q}\xspace}
\newcommand{\wocq}{\Delta \varepsilon_q\xspace}

\newcommand{\ext}{\textit{Ext.\%}}

\newcommand{\ci}{CI_{95\%}}

\definecolor{darkblue}{RGB}{47,85,151}
\definecolor{darkred}{RGB}{190,0,0}

\definecolor{darkpurple}{RGB}{153,0,204}
\definecolor{lightpurple}{RGB}{210,77,255}

\definecolor{lightergreen}{RGB}{115,181,74}
\definecolor{lightgreen}{RGB}{0,153,0}
\definecolor{darkgreen}{RGB}{0,102,0}

\iclrfinalcopy 
\begin{document}

\maketitle

\begin{abstract}
Human groups are able to converge on more accurate beliefs through deliberation, even in the presence of polarization and partisan bias --- a phenomenon known as the ``wisdom of partisan crowds.''  
Generated agents powered by Large Language Models (LLMs) are increasingly used to simulate human collective behavior, yet few benchmarks exist for evaluating their dynamics against the behavior of human groups. 
In this paper, we examine the extent to which the wisdom of partisan crowds emerges in groups of LLM-based agents that are prompted to role-play as partisan personas (e.g., Democrat or Republican). 
We find that they not only display human-like partisan biases, but also converge to more accurate beliefs through deliberation as humans do. 
We then identify several factors that interfere with convergence, including the use of chain-of-thought prompt and lack of details in personas. 
Conversely, fine-tuning on human data appears to enhance convergence.
These findings show the potential and limitations of LLM-based agents as a model of human collective intelligence.
\end{abstract}

\section{Introduction}

As Large Language Models (LLMs) have grown more human-like in the behaviors they produce \citep{park2022social}, there has been increasing interest in investigating whether they can be used to better understand emulate human communication in social groups \citep{tornberg2023simulating,kaiya2023lyfe,li2023quantifying,chuang2023simulating}. As one example, 
\citet{park2023generative} used LLMs to construct \textit{generative agents} that interact with each other in a simulated environment: initiating conversations, spreading information, remembering past events, and planning future actions. The resulting vignettes can show remarkably convincing interactions in which information about novel events, such as the planning of a birthday party, diffuses throughout the community of simulated agents. Yet it is difficult to understand how human-like such patterns really are, and consequently how useful such simulated systems are for understanding human communicative phenomena, without replicable empirical benchmarks of human behavior for comparison.



\begin{figure*}[t!] 
\centering
\includegraphics[width=0.99\linewidth]{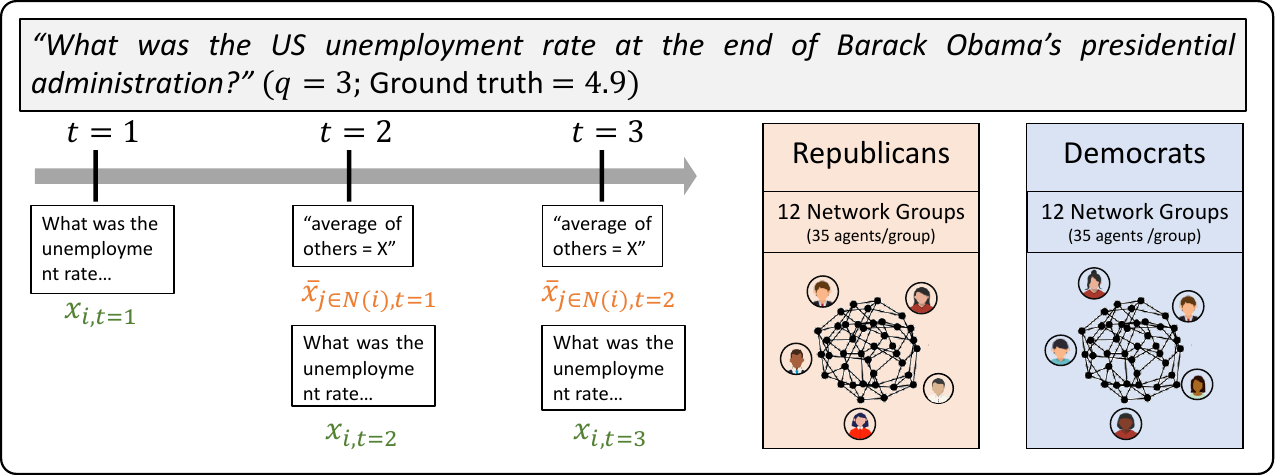}
\vspace{-2mm}
\caption{Experimental design comparing social feedback effects on LLM agents' estimations of partisan-biased factual questions. LLM agents role-playing \textcolor{darkblue}{Democrat} and \textcolor{darkred}{Republican} update their estimates after considering their peers' average responses \citep{becker2019wisdom}.}
\label{fig:exp_design_schematic}
\vspace{-4mm}
\end{figure*}

The current paper develops one such benchmark deriving from a phenomenon in the study of human collective intelligence, the \textit{wisdom of (partisan) crowds}. The phenomenon reflects two interesting characteristics of human cognition. First, when estimating real-world quantities related to a politically polarized issue, self-identified Republicans and Democrats often generate systematically different guesses that reflect their political leanings. For example, when asked to estimate the US employment rate during Barack Obama's administration, both groups overestimate relative to the ground truth, but Republicans produce reliably higher over-estimates, presumably remembering unemployment as worse than it actually was under the Democratic regime. 

Second, when shown the mean guess from other members of their preferred party, \textit{both} groups adjust their estimates in ways that move the group mean systematically closer to the ground truth. This phenomenon, known as the \emph{wisdom of crowds}, is a paradigmatic example of how groups pool individual knowledge through aggregation and deliberation \citep{kameda2022information,yi2012wisdom}. 
Moreover, when individuals are shown the average estimate of their group and allowed to adjust their own, the group's average becomes \textit{more accurate} \citep{becker2017network}, even for biased groups \citep{ becker2019wisdom}.
The wisdom of crowds effect where social influence improves collective estimates, extends across different cultures \citep{jayles2017social}, and finds application in practical domains such as clinical decision-making \citep{centola2023experimental} and science communication \citep{guilbeault2018social}. 

In a seminal study, \citet{becker2019wisdom} collected data from 1120 participants identifying as either Republican or Democrat. Each was asked factual questions known to elicit partisan bias, and after responding, was shown the average belief of others in their same partisan group (i.e. other Democrats or Republicans). Participants were then permitted to adjust their estimate, and the same procedure was repeated, yielding a series of three estimates for each respondent. The authors observed that, after each round of feedback, the mean estimate for both groups moved closer to the ground truth. 

This wisdom of partisan crowd phenomenon is useful for assessing LLM simulation of human communication for three reasons. First, all questions have a ground-truth value, providing a means of quantifying how accurate the actual LLM estimates are. 
Second, humans typically show partisan lean in their estimates. This provides an opportunity to evaluate whether role-playing LLMs show human-like patterns of partisan bias in their responses. Third, the social exchange of information within human partisan groups increased mean accuracy for each while also reducing polarization between groups, providing a reliable dynamic phenomenon in human communication that can be assessed in LLM agents. 

For these reasons, we replicated the experimental design of \citet{becker2019wisdom}, applying it to groups of interacting, role-playing LLM agents in a simulated environment, and assessing whether the resulting system replicated the key phenomena in human behavior. We found that LLM agents, when operating without Chain-of-Thought (CoT) reasoning, exhibit a substantial Wisdom of Partisan Crowds (WOC) effect, closely paralleling human patterns of error reduction in group settings. However, the use of CoT reasoning reduced this effect. We also show that the ``depth of persona'' created in the role-playing prompt critically influences whether LLM agents exhibit human-like partisanship bias in their estimates. Finally, fine-tuning LLMs with human data enhances human-like group dynamics in held-out data, though such training also risks overfitting. Together the work suggests a promising approach toward using established behavioral phenomena from human participants to evaluate and shape the use of LLMs for understanding dynamics of social communication.


\begin{table*}[bht!]
    \centering
    \small
    \caption{Evaluation of resemblance between LLM agent and human in social interaction setting. The three main human-LLM alignment metrics are, $HLI$ (the more positive, the more human-like,  $\woc$ (the more negative, the stronger the WOC effect) and $\pb$ (the more positive, the more aligned with human). The black boldface highlights the condition with the highest $HLI$. The metrics are shown with the standard errors. Notably, when there is \textcolor{lightgreen}{\textbf{no CoT}} reasoning, $\woc$ is always more negative than using \textcolor{lightergreen}{\textbf{CoT}} reasoning. In addition, using a \textcolor{darkpurple}{\textbf{detailed persona}} always leads to a more positive $\pb$ than using a \textcolor{lightpurple}{\textbf{simple persona}}. }    
    \resizebox{0.8\linewidth}{!}{  
    \begin{tabular}{@{}lllrrrl@{}}
        \toprule
        Model & Persona & CoT & $HLI \uparrow$ & $\woc \downarrow$ & $\pb \uparrow$ & $\ext$ \\ 
        \midrule
        ChatGPT & Detailed & CoT & 4.45 ± 0.8 & \textcolor{lightergreen}{\textbf{-1.08 ± 0.76}} & \textcolor{darkpurple}{\textbf{3.37 ± 0.25}} & 0.00 \\
         & & No CoT & \textbf{12.82 ± 1.89} & \textcolor{lightgreen}{\textbf{-7.59 ± 1.87}} & \textcolor{darkpurple}{\textbf{5.23 ± 0.28}} & 0.00 \\
         & Simple & CoT & -20.13 ± 1.1 & \textcolor{lightergreen}{\textbf{-2.07 ± 0.87}} & \textcolor{lightpurple}{\textbf{-22.2 ± 0.67}} & 0.00 \\
         & & No CoT & -21.8 ± 1.77 & \textcolor{lightgreen}{\textbf{-3.11 ± 1.47}} & \textcolor{lightpurple}{\textbf{-24.91 ± 0.98}} & 0.00 \\
        \cmidrule(r{4pt}){2-6} \cmidrule(l{4pt}){6-7}
        Vicuna-33B & Detailed & CoT & 2.81 ± 1.36 & \textcolor{lightergreen}{\textbf{2.87 ± 1.27}} & \textcolor{darkpurple}{\textbf{5.68 ± 0.49}} & 1.31 \\
         & & No CoT & \textbf{4.35 ± 2.64} & \textcolor{lightgreen}{\textbf{-0.68 ± 2.51}} & \textcolor{darkpurple}{\textbf{4.36 ± 0.80 }} & 1.38 \\
         & Simple & CoT & 3.36 ± 1.25 & \textcolor{lightergreen}{\textbf{0.59 ± 1.18}} & \textcolor{lightpurple}{\textbf{3.94 ± 0.41}} & 0.98 \\
         & & No CoT & -0.63 ± 2.63 & \textcolor{lightgreen}{\textbf{0.49 ± 2.47}} & \textcolor{lightpurple}{\textbf{-0.14 ± 0.91}} & 5.60 \\
        \midrule
         Human & - & - & 66.5 ± 6.79 & -33.16 ± 6.74 & 33.35 ± 0.83 & 8.37\\ 
        \bottomrule
    \end{tabular}
}    
    \label{tab:result_main}
\end{table*}

\section{Methods}\label{sec:methods}

\subsection{Experimental Procedure}
We followed the experimental design from \citet{becker2019wisdom}, using LLMs to role-play Democrat and  Republican personas. Each LLM agent is embedded in a network structure that governs interactions, connecting to four others sharing the same political leaning (all agents have node degree $k=4$; see Figure~\ref{fig:exp_design_schematic}) and thus reflecting the homogeneous group structures in human studies. Each LLM agent, powered by LangChain \citep{langchain} with OpenAI's ChatGPT (\texttt{gpt-3.5-turbo}; \citealp{openaiIntroducingChatGPT}) and the open-source LLM Vicuna (\texttt{vicuna-33B-v1.3}; \citealp{zheng2023judging}), maintains the continuity of each persona's memory throughout the experiment. Over three rounds, these agents were prompted to answer the same eight fact-based questions with known partisan biases as shown in Figure \ref{fig:exp_design_schematic}.  After each round, agents were given the average estimates of their connected peers and were asked to provide their estimates again. Thus at the end of the three rounds, each agent had produced three estimates for each of the eight questions.  \footnote{The original study by \citet{becker2019wisdom} divided these questions into two separate experiments, with the first four questions belonging to the first experiment and the last four to the second. However, as both sets of questions follow the same experimental procedure, we have merged them into a single analysis in our study.} The full list of questions is provided in \ref{app:list_questions}. The full procedure was conducted 12 times for each group to mirror the 12 groups of human participants in the social conditions in \citet{becker2019wisdom}. Temperate sampling (temperature = 0.7) was used to allow variability in responses. \ref{app:list_prompt} shows the actual prompts. The compute resource for using Vicuna is in \ref{app:compute_resource}.

\paragraph{Formal notation}
We denote each agent in the experiment as $a_{i,p,r}$, where $1 \leq i \leq 35$ indexes the agent within a specific run, $p$ denotes political leaning (Democrat or Republican; abbreviated as Dem and Rep hereafter), and $r$ specifies the run index with $1 \leq r \leq 12$. When the context is clear, we drop the subscript $p$ and $r$. For each political leaning, during each run, the agents answer eight questions over three time steps, generating a series of estimates $x_{i,q}^t$ for question $q$ at time $t$. Since all eight questions are fact-based, each has a ground truth value, denoted as $x^{*}_{q}$. Starting at $t \geq 2$, agents are shown $m^t_{i,q}$, the average estimate of their four politically homogeneous neighbors, before making their own estimates. 
\footnote{Formally, the average estimate from neighbors for agent $a_{i,p,r}$ at time $t$ for question $q$ is $m_{i,p,r,q}^t = \frac{1}{K} \sum_{j \in \mathcal{N}(i,p,r)} x_{j,p,r,q}^{t-1}$, where $\mathcal{N}(i,p,r)$ is the set of indices for the agents' neighbors who share the same political leaning $p$. The number of neighbors $K=4$.}


\subsection{Personas and Agent Specification}


\paragraph{Personas and agent specification} We prompted the LLMs to role-play as different personas created with varying levels of background detail. \textit{Simple Personas} are specified as ``a typical Democrat/Republican,'' relying on temperature sampling to elicit slightly different biased views. \textit{Detailed Personas} are provided with comprehensive backstories, including demographics and personal background information, to introduce individual differences based on such factors. 
This persona is retained in memory across the three rounds of adjustment for all questions. 


\ref{app:list_personas} shows the full list of both detailed and simple personas. A diverse set of detailed personas was generated by GPT-4. 

For example, \begin{quote}
    \textrm{Name: Isabella Johnson; Political leaning: Strong Democrat; Age: 67; Gender: Female; Ethnicity: White; Education: Bachelor's Degree in Education; Occupation: Retired Teacher; Background: Isabella is from Portland, Oregon, and spent her career advocating for public education and teachers' rights. She is passionate about social justice, healthcare, and environmental issues. Isabella is widowed with two grown children and enjoys birdwatching and painting in her free time.}
\end{quote}


\paragraph{Chain-of-thought reasoning (CoT)} 
We manipulated whether the agents used chain-of-thought (CoT) reasoning \cite{wei2022chain,wei2022emergent}. 
CoT has demonstrated success as a prompting strategy in solving complex reasoning tasks, such as arithmetic problems. However, recent work also indicates that CoT reasoning
may lead to stereotypes and biases~\cite{shaikh2022second}. This leads us to explore how CoT reasoning influences an LLM agent's ability to assume human-like behaviors in a social interaction setting. To elicit CoT reasoning, we append the prompt with the following: 
\begin{quote}
    ``Please provide your step-by-step reasoning and then give your estimate as a real number.''
\end{quote}

In contrast, in the condition without CoT reasoning, we end the prompt with 
\begin{quote}
    ``Please provide your estimate in a real number.''
\end{quote}


\subsection{Fine-Tuning the LLMs with Human Data}

In addition to in-context learning through prompting, we also perform supervised fine-tuning using human response data from the experiment in \citet{becker2019wisdom} to enhance the resemblance of human behaviors in LLM agents. Our fine-tuning methodology is inspired by \citet{binz2023turning}, showing that through supervised learning, LLMs can be adapted to modeling human decision-making behavior in an unseen task. We aim to investigate whether fine-tuning also improves the resemblance of human-like behavior in group interaction settings. We fine-tune two separate LLMs: one for Democrats and another one for Republicans. The training data consists of responses to questions $5 \leq q \leq 8$, while using questions $1 \leq q \leq 4$ as the testing set. The fine-tuned model is then evaluated separately on the train set and the test set. When fine-tuning, no persona is provided. Details on how we fine-tuned the LLM (ChatGPT in specific) are in the \ref{app:fine_tuning}.

\subsection{Evaluation Metrics}
\label{sec:eval_metrics}

\paragraph{Wisdom of Partisan Crowds Effect (WOC)} The Wisdom of Crowds effect quantifies the improvement in LLM agent estimates through social interaction, similar to that of human groups \cite{becker2019wisdom}. Within each political leaning and run, we compute the group mean for each question $q$ and time step $t$, $\bar{x}_{q}^{t} = \frac{1}{N} \sum_{i=1}^{N} x_{i,q}^t$ (with $N=35$ per group), and the normalized group mean $\eta_{q}^{t} = 100 \times (\bar{x}_{q}^t - x^{*}_{q})/{x^{*}_q}$. The normalized group error $\varepsilon_{q}^t = |\eta_{q}^{t}|$ shows the percentage deviation from the ground truth $|x^{*}|$. We measure the reduction in group error per question as $\Delta \varepsilon_{q} = \varepsilon_{q}^{t=3} - \varepsilon_{q}^{t=1}$, and average these across all questions, both political leanings, and all runs to obtain the average reduction in group error $\woc$. A more negative $\woc$ indicates a stronger wisdom of crowd effect, with $\woc$ representing the percentage of ground truth size $|x^{*}|$ by which estimates move toward truth. For detailed derivation, see \ref{app:metric_reduction_in_group_error}.

\paragraph{Partisan Bias} 

\begin{wrapfigure}[28]{R}{0.5\linewidth}
\centering
\includegraphics[width=0.99\linewidth]{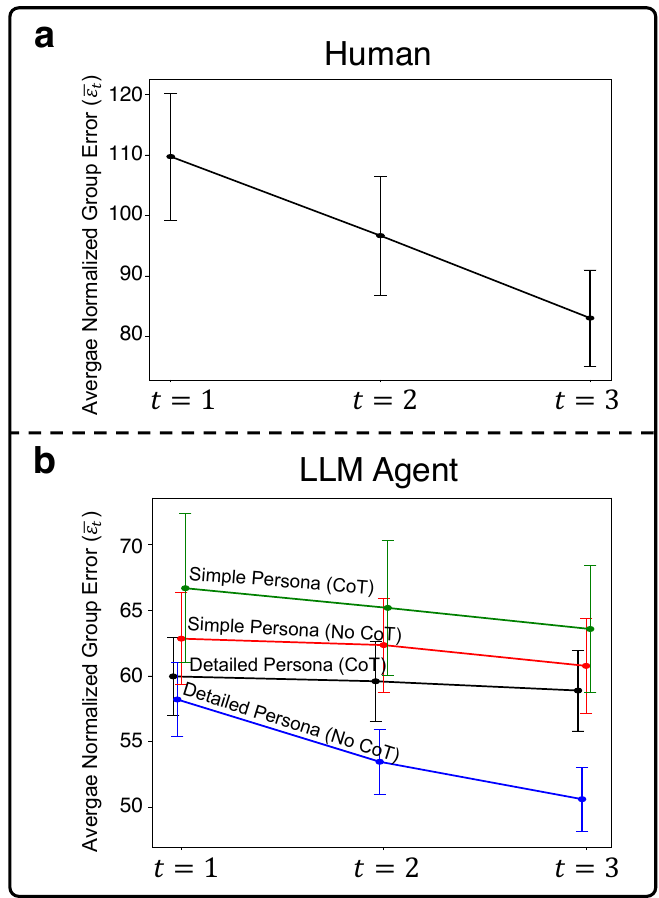}
\caption{Average Normalized Group Error ($\overline{\varepsilon}_t$) for (a) human crowds and (b) LLM agents (ChatGPT) across the experimental settings. Error bars indicating standard errors.}
\label{fig:becker_human_vs_llm}
\end{wrapfigure}

To evaluate human-like partisan biases in LLM agents, we define \textit{Partisan Bias} as the average difference in normalized group mean $\eta_{q}^t$ between the Democratic and Republican groups, in line with the expected directions of human partisan bias. Formally, for each questions $q$, let $\overline{D}_q$ be the normalized group mean $\eta_{q}^t$ averaged across Democrats' runs and time steps, and let $\overline{R}_q$ be the average for Republicans'. The partisan bias for question $q$ is defined as $\pbq = (\overline{R}_q - \overline{D}_q) \times \text{sign}(h_{q})$, where $\text{sign}(h_{q})$ indicates the human partisan bias direction as per human data \cite{becker2019wisdom}, with $+1$ if Republicans typically have greater $\eta_{p,r,q}^t$ than Democrats (i.e., a more positive $\overline{x}_{q}^t$ if $x^*_q>0$), $-1$ if the other way around, and $0$ if there is no expected difference. \footnote{$\text{sign}(h_{q})$: $+1$ in questions 3 (unemployment rate), 4 (taxes); $-1$ in questions 5 (military), 6 (immigration change), 7 (unemployment change); and $0$ in questions 1 (election), 2 (California), 8 (Soldiers).} In addition, we denote \textit{overall partisan bias} $\pb$ as the partisan bias averaged across all questions' $\pbq$. A positive $\pb$ indicates a overall similarity to the direction of human bias, and vice versa\footnote{Because $\eta_{q}^t$ is scaled by a factor of 100, $\pb$ can be interpreted as the partisan bias expressed in \textit{percentage} of the size of ground truth $|x^{*}|$}. 

\paragraph{Human Likeness Index} We introduce the Human Likeness Index (HLI) to assess the extent of LLM agents' resemblance to human behaviors. To aggregate the wisdom of crowd effect ($\woc$) and the partisan bias ($\pb$), we define $ HLI = \pb + (-\woc)$. A higher HLI score\footnote{A linear addition makes sense because both $\pb$ and $\woc$ are on the same scale. Both can be expressed as a percentage of the size of the ground truth value $|x^*|$.} indicates a stronger overall human-like behavior in the LLM agents within this group experiment.

\paragraph{Extreme Values ($\ext$)} The \textit{Ext.\%} metric evaluates the proportion of LLM agent responses that are unrealistic, based on established criteria \cite{becker2019wisdom}. For a fair comparison with human data, the same criteria are applied to identify extreme values, for example, marking any response to the unemployment rate above 47\% as extreme. These criteria are detailed in \ref{app:extreme_criteria}. Extreme values are excluded from calculations of \textit{Average Group Error Reduction} ($\woc$) and \textit{Partisan Bias} ($\pb$). The \textit{Ext.\%} thereby serves as a measure of the tendency of the LLM agents to generate unrealistic responses.

\paragraph{Revision Coefficient} In human crowds, the \textit{mechanism} for why the group mean converges towards the truth through social interaction is that those who are more accurate at their initial estimate tend to be influenced less by the information they received, and thus pull the group distribution towards the truth \citep{becker2017network}. Following \citet{becker2017network}'s methodology, for each question $q$, we calculate the \textit{revision coefficient} ($r_{\text{adj},q}$), defined as the partial correlation between \textit{individual revision} ($\Delta x_{i,q} = |x_{i,q}^{t=3} - x_{i,q}^{t=1}|$) and \textit{individual initial error} ($e_{i,q} = |x_{i,q}^{t=1} - x_{q}^*|$), adjusted for the \textit{social signal} ($s_{i,q} = |x_{i,q}^{t=1} - m_{i,q}^{t=2}|$) that each individual receives.  Adjusting for the social signal is important as individuals with higher initial errors often receive stronger social feedback as they deviate from the rest. Formally, $r_{\text{adj},q}=\text{corr}(\widetilde{\Delta x}_{i,q},\widetilde{e}_{i,q})$, where $\widetilde{\Delta x}_{i,q}$ and $\widetilde{e}_{i,q}$ are ${\Delta x}_{i,q}$ and ${e}_{i,q}$ adjusted by social signal. Please refer to \ref{app:revision_coefficient} for detailed derivation.


\section{Results and Discussion}\label{sec:results}

\begin{figure*}[bt!] 
\centering
\includegraphics[width=1\linewidth]{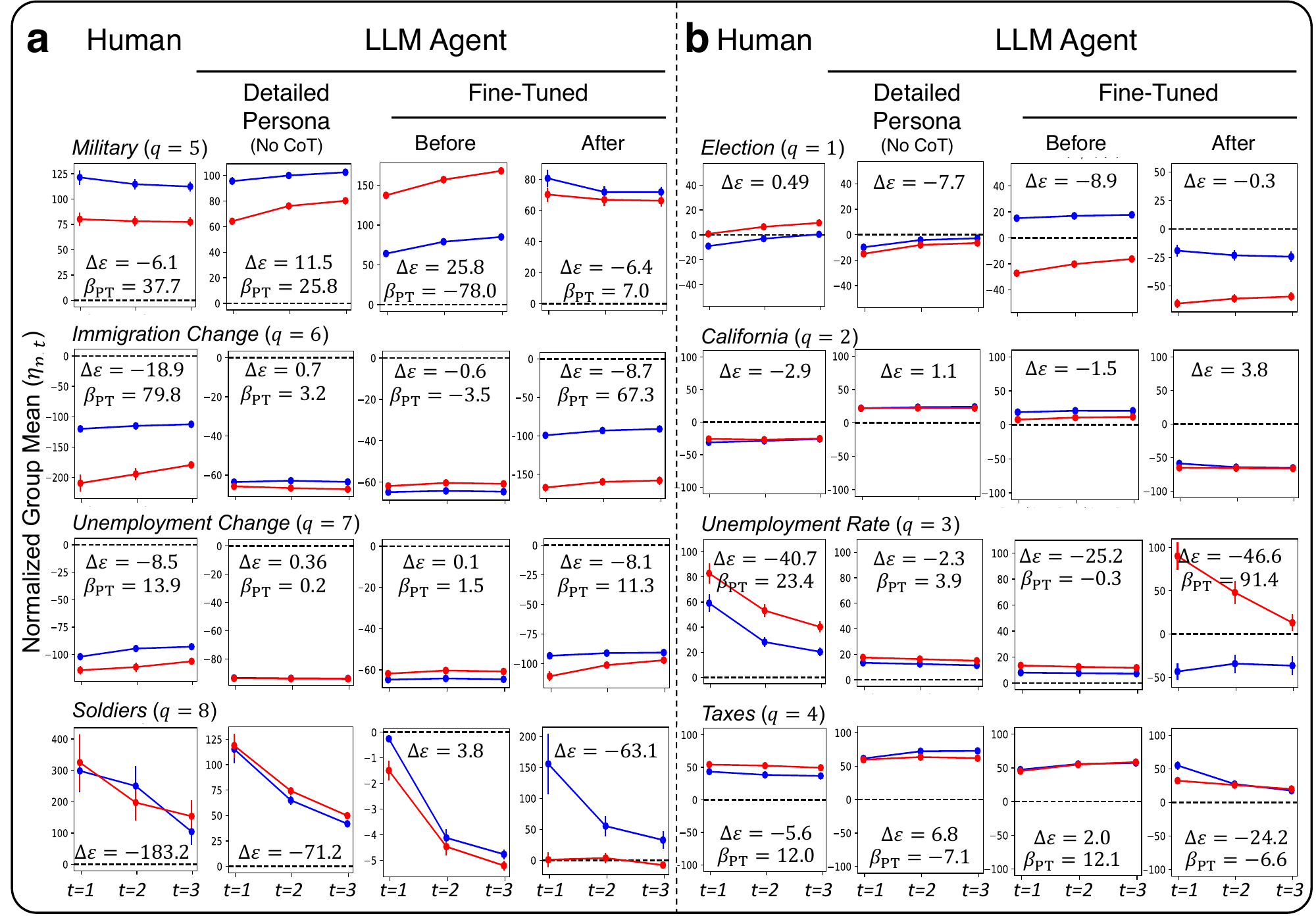}
\vspace{-2mm}
\caption{Normalized group mean $\eta_{p,t}$ over three rounds, averaged across 12 group experiments (\textcolor{red}{red} for Republicans, \textcolor{blue}{blue} for Democrats), with error bars for standard errors. Each panel consists of four columns representing different data sets: Column 1 shows human data. Columns 2 to 4 shows LLM (ChatGPT) agents' data. Column 2 depicts LLM role-playing detailed personas and without CoT reasoning (the configuration with the highest $HLI$); Column 3 presents LLM results before fine-tuning; and Column 4 illustrates LLM after fine-tuning. Panel (a) includes questions from the training set ($5 \leq q \leq 8$) used for fine-tuning the LLM agents, while Panel (b) displays questions from the hold-out test set ($1 \leq q \leq 4$). Question-specific WOC effects ($\wocq$) and partisan biases ($\pbq$, if expected) are overlaid for comparison.}
\label{fig:becker_human_vs_llm_full_list}
\vspace{-4mm}
\end{figure*}

\subsection{Effect of Persona Detail, CoT Reasoning}


\paragraph{Detailed Persona and without CoT Reasoning Elicits Wisdom of Crowds Effect}
LLM agents with detailed personas, but without CoT reasoning, demonstrate the closest resemblance to human group dynamics. They demonstrate the highest human likeness, $HLI=12.82$ (ChatGPT) and $3.67$ (Vicuna) among the six experimental settings (Table~\ref{tab:result_main}). Figure~\ref{fig:becker_human_vs_llm} visualizes the wisdom of partisan crowd results of LLM agents (ChatGPT). These agents converge significantly towards the ground truth after social interaction, quantified by a significant WOC effect, $\woc=-7.59$, $\ci=[-11.08,-4.10]$, $p<.001$. It also shows a significant human-like partisan bias, $\pb=5.23, \ci=[4.66,5.81]$, $p<.001$. \footnote{The p-values and 95\% Confidence Intervals ($\ci$) are derived from bootstrapping with 1000 resamplings \citep{efron1992bootstrap}.} Figure~\ref{fig:becker_human_vs_llm_full_list} shows the detailed result for each question. 

Next, we look into the role of persona detail and CoT reasoning, respectively. The result with Vicuna is shown in Figure~\ref{fig:becker_human_vs_llm_full_list_vicuna} in \ref{app:result_vicuna}. As shown in Table~\ref{tab:result_main}, without CoT reasoning, the agents' error reduction through social interaction is consistently greater than with CoT reasoning, $\woc$ (without CoT) $<$ $\woc$ (with CoT), difference $=4.63$, $\ci = [2.10,7,20]$, $p<.001$. For example, when role-playing detailed persona, the LLM agents (ChatGPT)' error reduction $\woc = -7.59$ when there is no CoT reasoning, as opposed to $\woc = -1.08$ with CoT reasoning, difference $=6.52$, $\ci = [2.59, 10.72]$, $p<.001$. In addition, as shown in Figure~\ref{fig:becker_human_vs_llm}b, without CoT reasoning always yield a smaller averaged normalized group error $\overline{\varepsilon}_t$ than the counterpart without CoT reasoning. The result with Vicuna shows similar patterns.



\paragraph{Detailed Persona and CoT Reasoning Encourages Human-like Partisan Bias}
The depth of persona detail and the use of CoT reasoning significantly increase the LLM agents' resemblance of human-like partisan bias $\pb$ (Table~\ref{tab:result_main}). Detailed personas enables a more human-like partisan bias across the two LLMs and across the two CoT reasoning conditions, $\pb$ (detailed persona) $>$ $\pb$ (simple persona), difference = $15.48$, $\ci=[14.63,16.36]$, $p<.001$. On the other hand, the use of CoT reasoning also enables a more human-like partisan bias across the two LLMs and across all conditions, $\pb$ (CoT) $>$ $\pb$ (no CoT), difference = $13.64$, $\ci=[12.48,14.78]$, $p<.001$.



This example illustrates how, in role-playing scenarios, LLM agents' CoT reasoning is more influenced by their personas than factual accuracy.

\subsection{Impact of Fine-Tuning on Enhancing Human-Like Dynamics}


As shown in Table~\ref{tab:result_fine_tune} and Figure~\ref{fig:becker_human_vs_llm_full_list}, in the training set (questions $5 \leq q \leq 8$), the human likeness index ($HLI$) increases to $50.11$ (from $-33.95$ before fine-tuning), partisan bias $\pb$ increases to $28.53$ from $-26.68$, difference $=55.20,\ci=[52.55,58.00],p<.001$, and the wisdom of crowds effect $\woc$ changes to $-21.59$ from $7.27$, difference $=28.86,\ci=[-41.52,-18.44],p<.001$. However, in the test set (questions $1 \leq q \leq 4$), there is an increase in extreme values ($\ext=29.94\%$), indicating a risk of overfitting. For example, the fine-tuned LLM agents tend to provide a negative estimate (up to $84.52\%$) for unemployment rate estimation ($q=3$), which is not valid and deemed as extreme values, presumably because there is a similarly worded question about the \textit{change} in unemployment rate where many humans provide negative estimates ($46.90\%$ of responses). Nonetheless, after filtering out extreme responses, the fine-tuned models continue to show strong human-like behavior, with an enhanced $HLI$ of $31.97$ (increased from $0.11$ before tuning), and $\pb=-14.1$ (increased from $2.31$, difference$=16.42,\ci=[-22.10,-10.17],p<.001$) and $\woc=-14.1$ (changed from $2.31$, difference$=15.67,\ci=[11.37,20.47],p<.001$). These findings suggest that fine-tuning can greatly enhance the human-like qualities of LLM agents, and even generalize well to unseen question if proper application of filtering criteria is applied.

\begin{table}[tb!]
    \small
    \centering
    \caption{Evaluation of fine-tuned LLM alignment with human group dynamics on the train set ($5 \leq q \leq 8$) and the test set ($1 \leq q \leq 4]$). The boldface highlights the consequences of fine-tuning. }    
        \begin{tabular}{@{}p{2cm}llll@{}} 
            \toprule
            Method & $HLI \uparrow$ & $\woc \downarrow$ & $\pb \uparrow$ & $\ext$ \\ 
            \midrule
            \multicolumn{2}{@{}l}{Before Fine-tuning} & & & \\ 
            \multicolumn{1}{@{\hspace{1em}}l}{Train} & -33.95 ± 1.58 & 7.27 ± 1.18 & -26.68 ± 1.04 & 0.00 \\
            \multicolumn{1}{@{\hspace{1em}}l}{Test} & -0.11 ± 0.75 & 2.31 ± 0.73 & 2.2 ± 0.14 & 0.00 \\         
            \midrule
            \multicolumn{2}{@{}l}{After Fine-tuning} & & & \\ 
            \multicolumn{1}{@{\hspace{1em}}l}{Train} & \textbf{50.11} ± 6.18 & -21.59 ± 6.12 & 28.53 ± 0.89 & 0.09 \\
            \multicolumn{1}{@{\hspace{1em}}l}{Test} & \textbf{31.97} ± 3.77 & -14.1 ± 3.02 & 17.87 ± 2.26 & \textbf{29.94} \\      
            \midrule
            Human & & & & \\ 
            \multicolumn{1}{@{\hspace{1em}}l}{Train} & 97.95 ± 13.02 & -54.15 ± 12.97 & 43.8 ± 1.20 & 8.11 \\ 
            \multicolumn{1}{@{\hspace{1em}}l}{Test} & 29.83 ± 2.21 & -12.16 ± 2.07 & 17.67 ± 0.78 & 8.64 \\             
            \bottomrule
        \end{tabular}
    \label{tab:result_fine_tune}
\end{table}

\begin{figure*}[tb!] 
\centering
\includegraphics[width=0.99\linewidth]{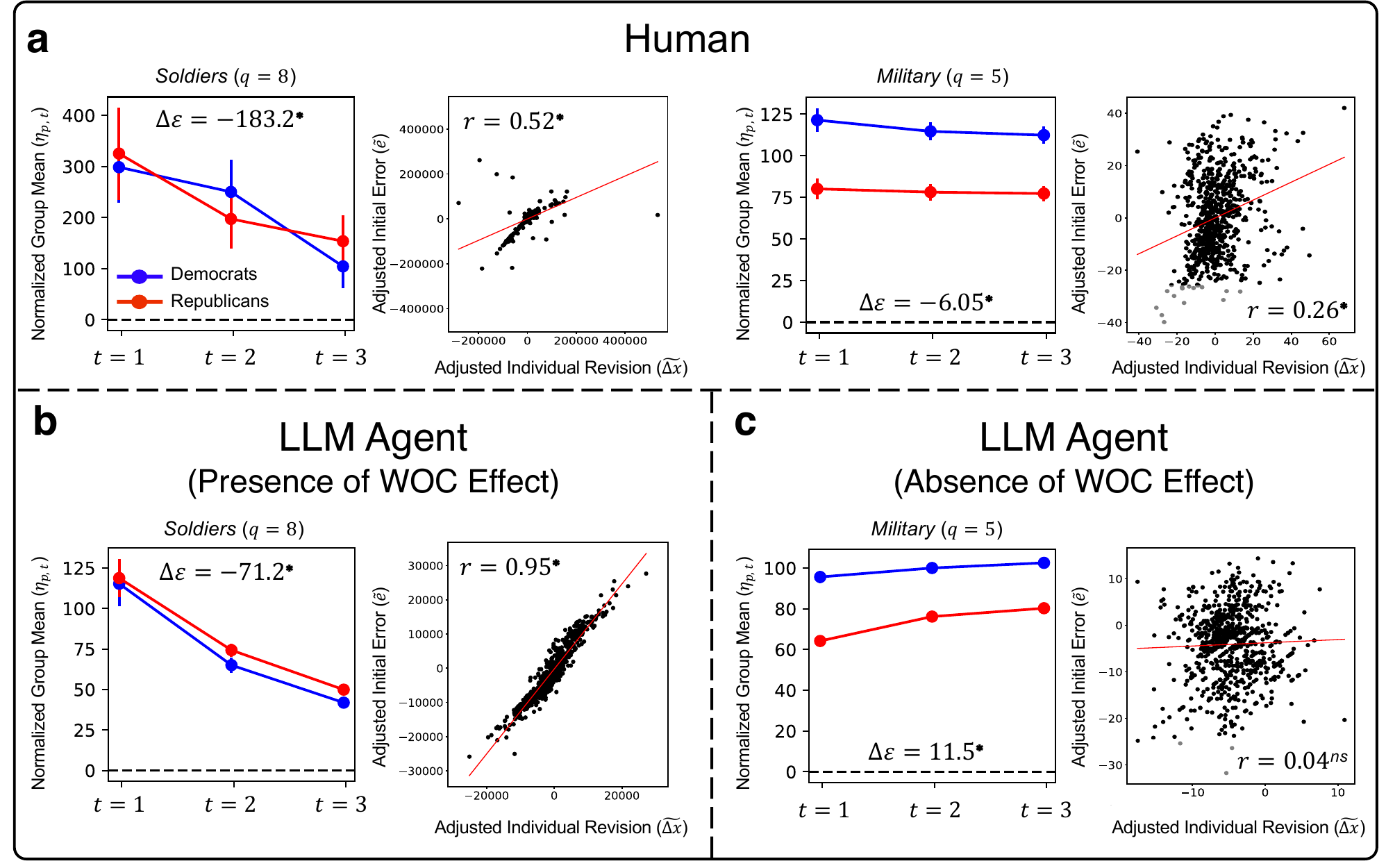}
\vspace{-2mm}
\caption{Mechanism of why the WOC effect emerges from crowds of Human crowds and LLM agents. Panel (a) and (b) shows the examples where both humans and LLM agents show the WOC effect  through social interaction (i.e., the question-specific WOC effect $\wocq < 0$). In contrast, in panel (c), LLM agents do not converge towards the ground truth while humans do. In each panel, the line plot shows the normalized group mean $\eta_{p,t}$ trajectory over three rounds, averaged across 12 runs (\textcolor{red}{red} for Republicans, \textcolor{blue}{blue} for Democrats), with error bars indicating standard errors. The $r$ in each panel demonstrate the revision coefficient - the correlation $r_\text{adj}$ between the adjusted initial individual error $\widetilde{e}_{i,p,r,q}$ and adjusted estimate revisions $\widetilde{\Delta x}_{i,p,r,q}$ (\ref{sec:eval_metrics}). Similar to human crowds, the LLM agents show the WOC effect only when $r_\text{adj}>0$. The results of the full set of questions are shown in Figure~\ref{fig:llm_align_misalign_full_list} (\ref{app:result_revision_coefficient}). \textbf{$^{*}$}: $p < .01$ (Bonferroni corrected for all questions); $^\textit{ns}$: not significant.}
\label{fig:human_vs_llm_align_misalign_AND_llm_align_misalign_short_list_fig}
\vspace{-4mm}
\end{figure*}


\subsection{Mechanism of the Wisdom of Crowds Effect}
\label{sec:result_revision_coefficient}

In human group dynamics, the wisdom of crowds (WOC) effect arises when individuals with initially accurate estimates are less influenced by others, as indicated by a \textit{positive revision correlation coefficient} \citep{becker2017network}. This mechanism is distinct from situations where error reduction is uniform across the group. Our analysis, detailed in Figure~\ref{fig:human_vs_llm_align_misalign_AND_llm_align_misalign_short_list_fig} and~\ref{fig:llm_align_misalign_full_list} reveals that LLM agents exhibit a similar pattern: the WOC effect occurs ($\wocq<0, ps<.001$\footnote{Bonferroni corrected for both the p values for both $\wocq < 0$ and $r_{adj} > 0$}) only when the revision coefficient is significantly positive ($r_{adj} > 0$, $ps<.001$). In contrast, when the revision coefficient is not positive, the WOC effect never emerges. In sum, LLM agents' WOC effect emerges through the same mechanism as human crowds, where those with more precise initial estimates exert greater influence on the group's final consensus.

\section{Related Work}\label{sec:related_work}

\paragraph{Simulating Human-like Group Dynamics with LLM-based Agents} 

Research in applying LLM-based agents for social simulation is expanding \citep{park2023generative,park2022social,kaiya2023lyfe,tornberg2023simulating, li2023quantifying}. However, these behaviors are not evaluated against actual human data and thus how human-like they are remains unclear. Similarly, \citet{tornberg2023simulating} simulate social media platforms using LLMs and agent-based modeling to evaluate the effect of different news feed algorithms, yet the comparison with human interactions is also absent. \citet{park2022social} demonstrates that LLM-driven agents can produce \textit{human-like} posts on platforms like Reddit that. However, their work doesn't test the agents' ability to represent demographic-specific behaviors such as political leanings. This highlights a critical gap in the field: evaluate the human likeness of demographic-based LLM agents in social interaction setting.

\paragraph{Wisdom of Crowds Through Social Interaction}
Research on the wisdom of crowds (WOC) effect has consistently shown that social interaction can refine group estimations in human crowds \citet{becker2017network} demonstrate that collective accuracy increases with information exchange in decentralized networks, and \citet{becker2019wisdom} show that it also manifests in politically homogeneous crowds. In addition, they identify the mechanism for why WOC emerges through social interaction: individuals with larger initial errors adjust their estimates more, contributing to group wisdom. In addition, the effect is robust across cultures \citep{jayles2017social}. It has also been applied to real-world application settings, such as clinical decision-making \citep{centola2023experimental} and science communication \citep{guilbeault2018social}. These studies validate the use of social information to improve collective intelligence as a robust benchmark to evaluate human-like behaviors in social interaction context.

\section{Conclusion}\label{sec:conclusion}

Our study utalize \citet{becker2019wisdom}'s experimental design to evaluate Large Language Models (LLMs)-based agents in a simulated environment. The findings shed light on their potential in emulating human group dynamics. We discover that LLM agents, when role-playing detailed personas, demonstrate a wisdom of partisan crowds effect, mirroring the error reduction seen in human groups. However, incorporating CoT reasoning or a lack of detailed persona tends to diminish this effect. Additionally, the level of detail in agents' personas significantly influences their display of human-like partisan biases. The fine-tuning of LLMs with human data further enhances their ability to replicate human-like group dynamics to unseen questions. This study highlights the potential of LLM-based agents in producing human-like group dynamics when grounded with empirical human data. 

Despite the artificial experimental setting \citep{becker2019wisdom}, our study point to a promising direction in using established behavioral phenomena of human participants to evaluate and shape LLMs for simulating human social communication dynamics. Looking forward, we envision that, by  incorporating human social interaction data into LLM agent development, future studies can develop human-emulating LLM agents for broader social simulations that have been traditionally tackled with agent-based models \citep{lorenz2021individual,flache2017models,chuang2023computational}.




\bibliography{main}
\bibliographystyle{iclr2024_conference}

\newpage
\appendix
\label{sec:appendix}

\section{Results of Revision Coefficient}
\label{app:result_revision_coefficient}

\begin{figure*}[htbp!] 
\centering
\includegraphics[width=0.99\linewidth]{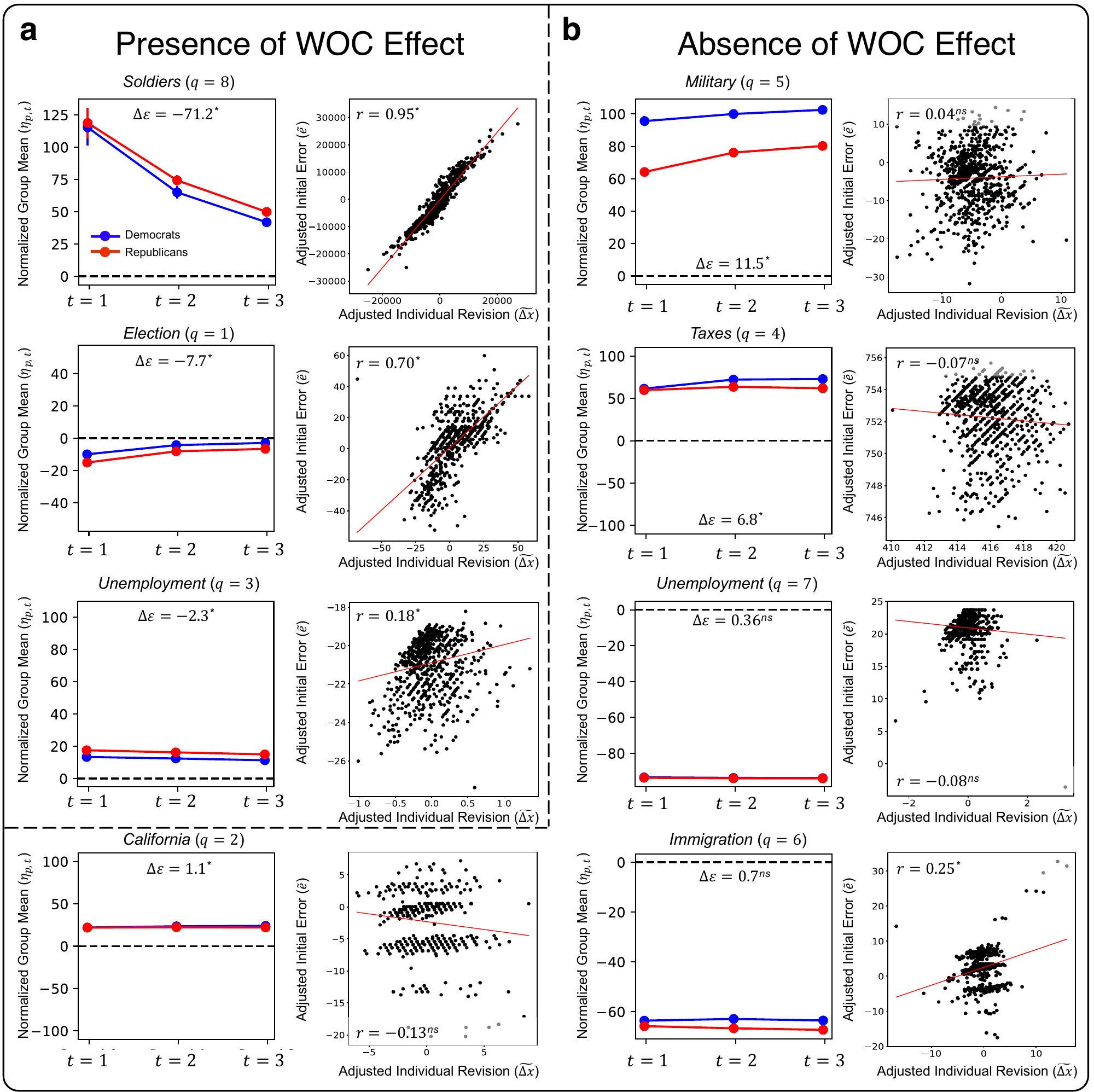}
\vspace{-2mm}
\caption{Analysis of the mechanism of LLM agents' wisdom of crowds (WOC) effect at the individual level. Panel (a) shows the questions where LLM agents exhibit the WOC effect ($\Delta \varepsilon_q < 0)$. Panel (b) shows the questions where LLM agents do not show the WOC effect. Within each panel, the questions are ordered by their revision correlation $r_\text{adj}$. In each panel, the line plot shows the normalized group mean $\eta_{p,t}$ trajectory over three rounds, averaged across 12 runs (\textcolor{red}{red} for Republicans, \textcolor{blue}{blue} for Democrats), with error bars indicating standard errors. The in each panel demonstrate the revision coefficient, the correlation $r_\text{adj}$ between the adjusted initial individual error $\widetilde{e}_{i,p,r,q}$ and adjusted estimate revisions $\widetilde{\Delta x}_{i,p,r,q}$ (\ref{sec:eval_metrics}). The LLM agents show the WOC effect only when $r_\text{adj}>0$ (panel a). \textbf{$^{*}$}: $p < .01$ (Bonferroni corrected for all questions); $^\textit{ns}$: not significant.}
\label{fig:llm_align_misalign_full_list}
\vspace{-4mm}
\end{figure*}

Figure~\ref{fig:llm_align_misalign_full_list} detailed in the analysis of revision coefficients. Specifically, the wisdom of crowds effect emerges only when the revision coefficient is positive, mirroring human behavior. Like human crowds, LLM agents enhanced collective accuracy when individuals with initially more accurate estimates are less influenced by peer opinions, thereby steering the group toward more accurate collective estimates. 

\section{Vicuna Results}\label{app:result_vicuna}

\begin{figure*}[bthp!] 
\centering
\includegraphics[width=0.7\linewidth]{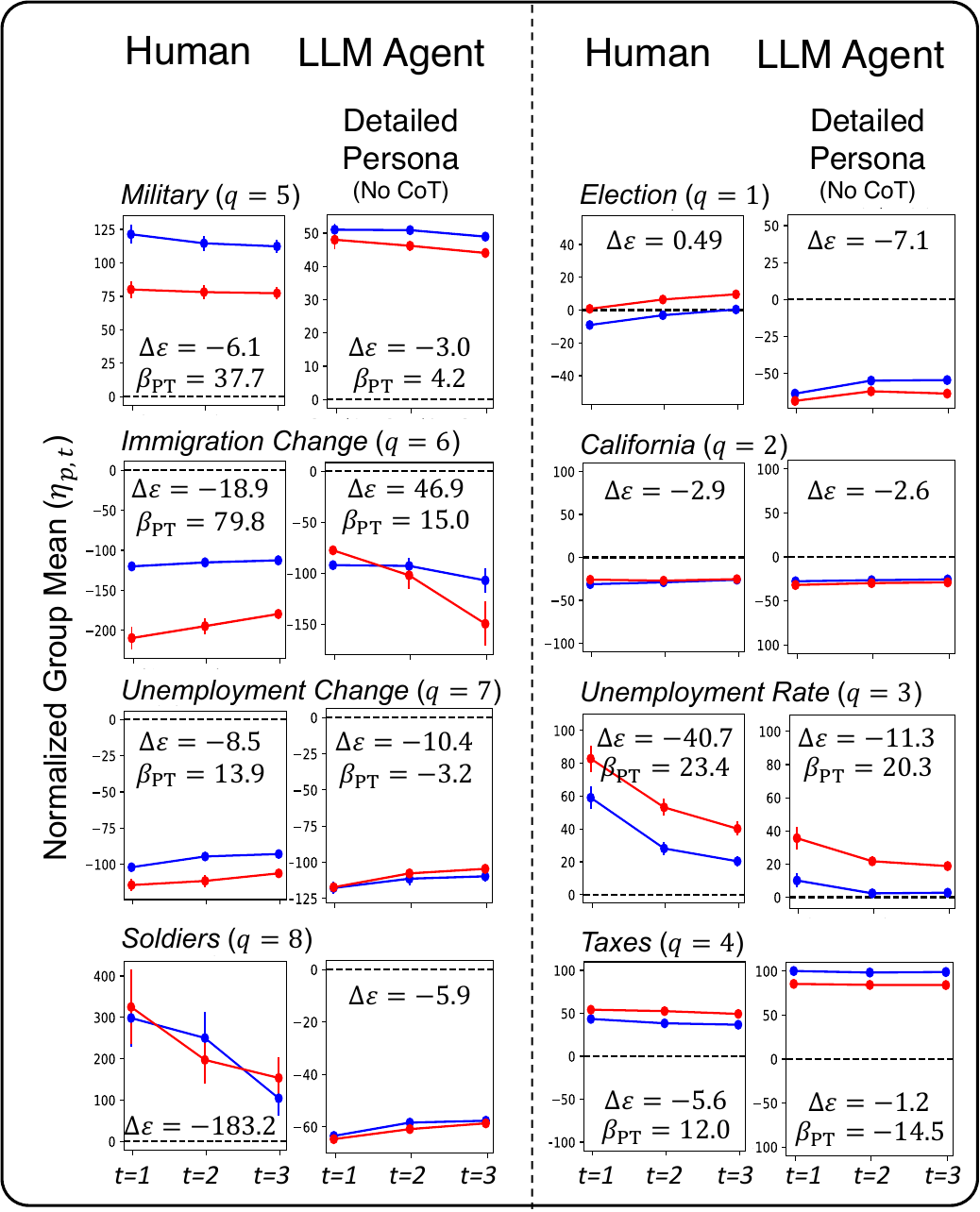}
\vspace{-2mm}
\caption{Trajectories of normalized group mean $\eta_{p,t}$ over three rounds, averaged across 12 group experiments (\textcolor{red}{red} for Republicans, \textcolor{blue}{blue} for Democrats), with error bars for standard errors. Each panel consists of two columns representing different data sets: Column 1 shows human data. Columns 2 shows LLM (Vicuna) agents' data. We separate questions into two panels to facilitate comparison with Figure~\ref{fig:becker_human_vs_llm_full_list}, where panel (a) includes questions $5 \leq q \leq 8$, and (b) displays questions $1 \leq q \leq 4$. Question-specific WOC effects ($\wocq$) and partisan biases ($\pbq$, if expected) are overlaid for comparison.}
\label{fig:becker_human_vs_llm_full_list_vicuna}
\vspace{-4mm}
\end{figure*}

Figure~\ref{fig:becker_human_vs_llm_full_list} shows the result for human versus ChatGPT-based LLM agents. Here, we shows the results for human versus Vicuna-based LLM agents in Figure~\ref{fig:becker_human_vs_llm_full_list_vicuna}. As shown, Vicuna-based LLM agents show the WOC effect ($\woc<0$) for all questions except the question $q=6$, $p<.001$ (Bonferroni corrected).

\section{List of Personas}\label{app:list_personas}

\subsection{Detailed Persona Condition}
In this section, we list the personas of the agents interacting in our experiment. 35 personas were used in the experiment for both Democrats and Republicans. For brevity purposes, a subset of them is included here. The full set of personas will be released along with the codebase at the time of publication. The list of personas is generated using this prompt ($\text{PARTY}$ is either Republican or Democrat) with GPT4 \citep{openaiIntroducingChatGPT}:

\begin{quote}
    
\tt \small Create detailed descriptions for 35 individuals who identify as {PARTY} and are above the age of 18, with varying degrees of political leaning (from Lean {PARTY} to Strong {PARTY}). Ensure that their demographics are diverse and reflective of the actual distribution of the US population, including factors such as age, gender, ethnicity, education level, and occupation. Provide comprehensive background information for each person. Use the format below.
Name:[]
Political leaning: []
Age: []
Gender: []
Ethnicity: []
Education: []
Occupation: []
Background: []
\end{quote}

\paragraph{Democrats}

\begin{mdframed}
\begin{quote}
    \tt \small\textbf{Name:} Isabella Johnson\\
    \textbf{Political leaning:} Strong Democrat\\
    \textbf{Age:} 67\\
    \textbf{Gender:} Female\\
    \textbf{Ethnicity:} White\\
    \textbf{Education:} Bachelor's Degree in Education\\
    \textbf{Occupation:} Retired Teacher
    \textbf{Background:} Isabella is from Portland, Oregon, and spent her career advocating for public education and teachers' rights. She is passionate about social justice, healthcare, and environmental issues. Isabella is widowed with two grown children and enjoys birdwatching and painting in her free time.
\end{quote}
\end{mdframed}

\begin{mdframed}
\begin{quote}
\tt \small\textbf{Name:} Jamal Brown\\
\textbf{Political leaning:} Lean Democrat\\
\textbf{Age:} 51\\
\textbf{Gender:} Male\\
\textbf{Ethnicity:} African American\\
\textbf{Education:} Bachelor's Degree in Finance\\
\textbf{Occupation:} Financial Analyst
\textbf{Background:} Jamal grew up in Detroit, Michigan, and became politically active during the 2008 recession. He supports policies promoting economic fairness and equal opportunities. Jamal is divorced with one child and enjoys playing golf and attending jazz concerts.
\end{quote}
\end{mdframed}

\begin{mdframed}
\begin{quote}
\tt \small\textbf{Name:} Karen Patel\\
\textbf{Political leaning:} Moderate Democrat\\
\textbf{Age:} 34\\
\textbf{Gender:} Female\\
\textbf{Ethnicity:} Indian American\\
\textbf{Education:} Master's Degree in Computer Science\\
\textbf{Occupation:} Software Engineer\\
\textbf{Background:} Karen was born in San Jose, California, and is a strong advocate for affordable housing and tech industry regulations. She also supports women's rights and STEM education. Karen is married with two young children and enjoys hiking and coding projects in her free time.
\end{quote}
\end{mdframed}

\begin{mdframed}
\begin{quote}
    \tt \small\textbf{Name:} Larry Jackson\\
    \textbf{Political leaning:} Strong Democrat\\
    \textbf{Age:} 42\\
    \textbf{Gender:} Male\\
    \textbf{Ethnicity:} White\\
    \textbf{Education:} Bachelor's Degree in Sociology\\
    \textbf{Occupation:} Nonprofit Fundraiser\\
    \textbf{Background:} Larry hails from Madison, Wisconsin, and is deeply involved in local politics. He is passionate about income inequality, racial justice, and LGBTQ+ rights. Larry is married with no children and enjoys traveling and volunteering for progressive causes.
\end{quote}
\end{mdframed}

\begin{mdframed}
\begin{quote}
    \tt \small\textbf{Name:} Monica Rodriguez\\
    \textbf{Political leaning:} Lean Democrat\\
    \textbf{Age:} 48\\
    \textbf{Gender:} Female\\
    \textbf{Ethnicity:} Puerto Rican\\
    \textbf{Education:} Associate Degree in Nursing\\
    \textbf{Occupation:} Registered Nurse\\
    \textbf{Background:} Monica grew up in New York City, New York, and supports policies that improve healthcare access and quality. She is also concerned about affordable housing and education reform. Monica is a single mother of two and enjoys salsa dancing and cooking.
\end{quote}
\end{mdframed}

\paragraph{Republicans}

\begin{mdframed}
\begin{quote}
    \tt \small\textbf{Name:} Charles Matthews\\
    \textbf{Political leaning:} Lean Republican\\
    \textbf{Age:} 38\\
    \textbf{Gender:} Male\\
    \textbf{Ethnicity:} African American\\
    \textbf{Education:} Bachelor's degree in Business Administration\\
    \textbf{Occupation:} Small business owner\\
    \textbf{Background:} Charles was born and raised in Atlanta, Georgia, where he attended a public university before starting his own business. He has a wife and two children. Charles supports limited government intervention and lower taxes, and he values entrepreneurship and self-reliance.
\end{quote}
\end{mdframed}

\begin{mdframed}
\begin{quote}
    \tt \small\textbf{Name:} Alice Thompson\\
    \textbf{Political leaning:} Moderate Republican\\
    \textbf{Age:} 29\\
    \textbf{Gender:} Female\\
    \textbf{Ethnicity:} Caucasian\\
    \textbf{Education:} Master's degree in Public Health\\
    \textbf{Occupation:} Epidemiologist\\
    \textbf{Background:} Alice grew up in a small town in Ohio before moving to Philadelphia for her studies. She is passionate about public health and believes in data-driven policies. She leans Republican due to her fiscal conservatism and support for individual rights.
\end{quote}
\end{mdframed}

\begin{mdframed}
\begin{quote}
    \tt \small\textbf{Name:} Juan Rodriguez\\
    \textbf{Political leaning:} Strong Republican\\
    \textbf{Age:} 45\\
    \textbf{Gender:} Male\\
    \textbf{Ethnicity:} Hispanic\\
    \textbf{Education:} High school diploma\\
    \textbf{Occupation:} Construction worker\\
    \textbf{Background:} Juan, originally from Mexico, migrated to Texas with his family when he was a child. A father of three, he believes in traditional family values, hard work, and limited government. He is a staunch advocate for securing the nation's borders.
\end{quote}
\end{mdframed}

\begin{mdframed}
\begin{quote}
    \tt \small\textbf{Name:} Sarah Chang\\
    \textbf{Political leaning:} Lean Republican\\
    \textbf{Age:} 23\\
    \textbf{Gender:} Female\\
    \textbf{Ethnicity:} Asian American\\
    \textbf{Education:} Bachelor's degree in Environmental Science\\
    \textbf{Occupation:} Environmental Consultant\\
    \textbf{Background:} Sarah was born and raised in California. She supports free-market solutions to environmental issues and believes in responsible resource management. Sarah leans Republican due to her fiscally conservative views and her opposition to excessive government regulation.
\end{quote}
\end{mdframed}

\begin{mdframed}
\begin{quote}
    \tt \small\textbf{Name:} Robert Klein\\
    \textbf{Political leaning:} Moderate Republican\\
    \textbf{Age:} 64\\
    \textbf{Gender:} Male\\
    \textbf{Ethnicity:} Caucasian\\
    \textbf{Education:} Bachelor's degree in Engineering\\
    \textbf{Occupation:} Retired engineer\\
    \textbf{Background:} Robert, a native of Pennsylvania, worked for a major manufacturing company for over 30 years. He is a Vietnam War veteran and a strong supporter of the Second Amendment. Robert believes in fiscal responsibility, limited government, and a strong national defense.
\end{quote}
\end{mdframed} 

\subsection{Simple Persona Condition}

\paragraph{Democrats}
\begin{mdframed}
\begin{quote}
\tt \small A typical Democrat in the USA.
\end{quote}
\end{mdframed}

\paragraph{Republicans}
\begin{mdframed}
\begin{quote}
\tt \small A typical Republican in the USA.
\end{quote}
\end{mdframed}
\section{Full List of Questions}\label{app:list_questions}

Below is the full list of questions ($q \in [1,8]$) in the experiment, along with the ground truth $x^*_q$ and the sign of human partisan bias direction $\text{sign}(h_q)$ observed in human data \citep{becker2019wisdom}.

\begin{enumerate}
    \item In the 2004 election, individuals gave \$269.8 million to Republican candidate George W. Bush. How much did they give to Democratic candidate John Kerry? (Answer in millions of dollars—e.g., 1 for \$1 million.) [$x_1^* = 224.6$, $\text{sign}(h_1) = 0$]
    \item According to 2010 estimates, what percentage of people in the state of California identify as Black/African-American, Hispanic, or Asian? (Give a number from 0 to 100.) [$x_2^* = 60.2$, $\text{sign}(h_2) = 0$]
    \item What was the US unemployment rate at the end of Barack Obama’s presidential administration—i.e., what percentage of people were unemployed in December 2016? (Give a number from 0 to 100.) [$x_3^* = 4.9$, $\text{sign}(h_3) = 1$]
    \item In 1980, tax revenue was 18.5\% of the economy (as a proportion of GDP). What was tax revenue as a percentage of the economy in 2010? (Give a number from 0 to 100.) [$x_4^*: 14.6$, $ \text{sign}(h_4) = 1$]

    \item For every dollar the federal government spent in fiscal year 2016, about how much went to the Department of Defense (US Military)? Answer with a number between 0 and 100. [$x_5^* = 15$, $ \text{sign}(h_5) = -1$]
    \item In 2007, it was estimated that 6.9 million unauthorized immigrants from Mexico lived in the United States. How much did this number change by 2016, before President Trump was elected? Express your answer as a percentage of change. [$x_6^* = -27.8$, $ \text{sign}(h_6) = -1$]
    \item How much did the unemployment rate in the United States change from the beginning to the end of Democratic President Barack Obama’s term in office? Express your answer as a percentage of change. [$x_7^* = -46$, $ \text{sign}(h_7) = -1$]
    \item About how many US soldiers were killed in Iraq between the invasion in 2003 and the withdrawal of troops in December 2011? [$x_78* = 4486$, $ \text{sign}(h_8) = 0$]
\end{enumerate}

\section{List of Induced Partisanship}\label{app:list_induced_biases}

Below is the full list of induced partisanship corresponding to the questions in the experiment ($q \in [1,8]$). We did not induce a bias for certain questions because \citep{becker2019wisdom} found that there is no significant human partisan bias for those questions. 


\begin{enumerate}
    \item Did not induce a bias for this question.
    \item Did not induce a bias for this question.
    \item Republicans tend to estimate a higher unemployment rate than democrats.
    \begin{enumerate}
        \item \textbf{Democrats:} "\textit{You believe that Barack Obama did a good job in reducing the US unemployment rate.}"
        \item \textbf{Republicans:} "\textit{You believe that Barack Obama did a poor job in reducing the US unemployment rate.}"
    \end{enumerate}
    \item Republicans tend to estimate higher tax revenue than democrats.
    \begin{enumerate}
        \item \textbf{Democrats:} "\textit{You believe that tax rates are not as high as they should be in general.}"
        \item \textbf{Republicans:} "\textit{You believe that tax rates are too high in general.}"\\
    \end{enumerate}

    \item Democrats tend to estimate a higher military budget than Republicans.
    \begin{enumerate}
        \item \textbf{Democrats:} "\textit{You believe that the US federal budget spent on the US military is too high in general.}"
        \item \textbf{Republicans:} "\textit{You believe that the US federal budget spent on the US military is not as high as it should be in general.}"
    \end{enumerate}
    \item Republicans tend to estimate higher number of immigrants than democrats
    \begin{enumerate}
        \item \textbf{Democrats:} "\textit{You believe that the unauthorized immigrants from Mexico were not a major national crisis before President Trump was elected.}"
        \item \textbf{Republicans:} "\textit{You believe that the US federal budget spent on the US military is not as high as it should be in general.}"
    \end{enumerate}
    \item Republicans tend to estimate a higher unemployment rate than democrats.
    \begin{enumerate}
        \item \textbf{Democrats:} "\textit{You believe that Barack Obama did a good job in reducing the US unemployment rate.}"
        \item \textbf{Republicans:} "\textit{You believe that Barack Obama did a poor job in reducing the US unemployment rate.}"
    \end{enumerate}
    \item Did not induce a bias for this question. \\
\end{enumerate}
\section{Full list of Prompts}\label{app:list_prompt}

In this section, we detail the prompts we use at each time step in the experiments.

\begin{mdframed}
\begin{quote}
    \tt \small
    \textbf{Round 1:} Role play this person.
\hl{\{AGENT\_PERSONA\}} \\
\hl{\{INDUCED\_BIAS\}}

Let's play a game where you'll be asked a single question, and you must provide an answer. This game has 3 trials, allowing you 2 chances to revise your response. Keep in mind that it's a group game, played concurrently with other participants. After you submit your first answer, you'll be given the average of other players' initial responses. Following your second submission, you'll receive the average of their second-round responses. At the end of the game, the more accurate your final answer is compared to the actual truth, the more money you will earn.

Now, \hl{\{AGENT\_NAME\}}, in this first round of the game, you are asked to answer the following question.

\hl{\{QUESTION\_CONTENT\}}

You must give an answer even if you are not sure.

Use the following format:

\hl{My Reasoning: [YOUR (\{AGENT\_NAME\}'s) STEP-BY-STEP REASONING]}

My Final Answer: [YOUR (\{AGENT\_NAME\}'s) ESTIMATE (A REAL NUMBER)]
\end{quote}
\end{mdframed}

\begin{mdframed}
\begin{quote}
\tt \small
    \textbf{Round 2:} Now, \{AGENT\_NAME\}, in this second round of the game, you are asked again to answer the following question.

\{QUESTION\_CONTENT\}

You must give an answer even if you are not sure.

This time, you are provided with other players in the first round of the game, who are all \hl{\{AGENT\_PARTY\}}. Their average answer: \hl{\{FEEDBACK\}}

Use the following format:

My Reasoning: [YOUR (\{AGENT\_NAME\}'s) STEP-BY-STEP REASONING]

My Final Answer: [YOUR (\{AGENT\_NAME\}'s) ESTIMATE (A REAL NUMBER)]
\end{quote}
\end{mdframed}

\begin{mdframed}
\begin{quote}
\tt \small
    \textbf{Round 3:} Now, \{AGENT\_NAME\}, in this third round of the game, you are asked again to answer the following question.

\{QUESTION\_CONTENT\}

You must give an answer even if you are not sure.

This time, you are provided with other players in the first round of the game, who are all \{AGENT\_PARTY\}. Their average answer: \{FEEDBACK\}

Use the following format:

\hl{My Reasoning: [YOUR (\{AGENT\_NAME\}'s) STEP-BY-STEP REASONING]}

My Final Answer: [YOUR (\{AGENT\_NAME\}'s) ESTIMATE (A REAL NUMBER)]
\end{quote}
\end{mdframed} 

In the prompt outline above, the highlighted sections are subject to change based on the configuration of the prompt:

\paragraph{AGENT\_PERSONA:}If we are using the detailed personas condition for a prompt, this placeholder is replaced with one of the descriptions from \ref{app:list_personas}.

\paragraph{INDUCED\_BIAS:}This placeholder is replaced with a bias corresponding to the party of the agent and the question. A list of these biases can be found in \ref{app:list_induced_biases}. 

\paragraph{AGENT\_NAME:} For the detailed persona condition, this placeholder is derived from the agent persona. For the simple persona condition, this is replaced with a generic name such as "\textit{r\_1}" for republicans and "\textit{d\_1}" for democrats.

\paragraph{My Reasoning  [YOUR (AGENT\_NAME's) STEP-BY-STEP REASONING]:} If the agent do not use chain-of-thought (CoT) reasoning in its responses, this section is excluded from the prompt at each time step. otherwise, it is included. 

\paragraph{AGENT\_PARTY:} This placeholder is replaced with either "\textit{democrats}" or "\textit{republicans}" depending on the agent's party affiliation.

\paragraph{FEEDBACK:} For the second and third round prompts, this placeholder is replaced with the mean response of the agent's neighbours for the previous round.

\paragraph{QUESTION\_CONTENT:} This placeholder is replaced with one of the question from \ref{app:list_questions}.

\section{Criteria for Extreme Values}\label{app:extreme_criteria}
For all questions in the experiment ($q\in[1,8]$), we use the same criteria as in the human study \citep{becker2019wisdom}\footnote{\citet{becker2019wisdom} use a log function to normalize the responses for $q \in [1,4]$ without removing extreme values. In our study, we do not do this since LLM agents may occasionally return negative values. Instead, we follow their ``alternative normalization procedire'' outlined in their supplementary materials. They show that different normalization procedures do not yield significant differences.}. The criteria are as follows:
\begin{enumerate}
    \item Election donations for John Kerry in 2004: Answers above \$2246 million ($10 \times x^*$) or below 0 are considered extreme.
    \item California demographics: Answers above 602\% ($10 \times x^*)$ or below 0\% are considered extreme.
    \item Unemployment rate at the end of Obama's presidency: Answers above 49\% ($10 \times x^*$) or below 0\% are considered extreme.
    \item Tax revenue as a percentage of GDP in 2010: Answers above 146\% ($10 \times x^*$) or below 0\% are considered extreme.
    \item Military spending: Answers above 100\% or below 0\% are considered extreme.
    \item Immigration rate changes: Answers above 1000\% or below -1000\% are extreme.        
    \item Unemployment rate changes:  Answers above 1000\% or below -1000\% are extreme.    
    \item Soldier deaths: Answers above 1 million and answers below 0 are extreme.    
\end{enumerate}

These criteria are based on the realistic ranges expected for these measures, as per the guidelines in \citet{becker2019wisdom}.

\section{Fine-tuning Details}\label{app:fine_tuning}

OpenAI's ChatGPT (\texttt{gpt-3.5-turbo-0613}) was fine-tuned using the human data from \citep{becker2019wisdom} to investigate the change in human-AI alignment in a group interaction setting. Two separate models were fine-tuned: one for Democrats and another one for Republicans.

These were the hyper-parameters used in the fine-tuning:
\begin{itemize}
    \item{Training Set Size:} 2747
    \item{Validation Set Size:} 381
    \item{Number of Epochs:} 4
    \item{Batch Size:} 5
    \item{Learning Rate Decay Factor:} 0.05
\end{itemize}


Human data was collected for 12 groups. For each agent $a_{i,p,r}$, a list of 4 neighbours $\mathcal{N}(i,p,r)$ was provided. We recreated the feedback provided to agents in rounds 2 and 3 by taking the average of their neighbours' estimates in the previous round, given by $m_{i,p,r,q}^t = \frac{1}{4} \sum_{j \in \mathcal{N}(i,p,r)} x_{j,p,r,q}^{t-1}$. The data for each agent was separated into 3 prompt-response pairs, corresponding to a specific round in the experiment. Each pair included the prompt and its corresponding response from the respective round, as well as those from earlier rounds. These pairs were passed into OpenAI's fine-tuning API and the experiment was run on the resulting models.

\section{Compute Resources}\label{app:compute_resource}
For running the Vicuna-33B model, we use a Linux server with 1 GPU card (A100-SXM4-80GB), taking about 12 hours to run the full experiment.
\section{Detailed Definition of Notations}

This appendix section elaborates on the specific definitions and the derivation process of key metrics and terms used throughout our study. 

\subsection{Reduction in Group Error Through Social Interaction}\label{app:metric_reduction_in_group_error}
The primary indicator of human-like performance in this group experiment setting is the ability to improve estimates over time through social interaction, as per \citep{becker2019wisdom}. To quantify this effect, we define several key terms:

\paragraph{Group Mean ($\bar{x}_{p,r,q}^t$):} For each run $r$ and each political leaning $p$, the group mean at each time step $t$ for question $q$ is denoted as $\bar{x}_{p,r,q}^t$. It represents the average of the estimates provided by all agents in the group at that time step and is expressed as $\bar{x}_{p,r,q}^t = \frac{1}{N} \sum_{i=1}^{N} x_{i,p,r,q}^t$, where $N=35$ is the total number of agents in the group.

\paragraph{Group Error ($\delta_{p,r,q}^t$):} The group error for each run $r$, political leaning $p$, and time step $t$ for question $q$ is denoted as $\delta_{p,r,q}^t$. This metric is calculated as the difference between the group mean and the ground truth value: $\delta_{p,r,q}^t = |\bar{x}_{p,r,q}^t - x^*_{q}|$. It represents the raw error of the group's collective estimate in comparison to the actual value for each question.

\paragraph{Normalized Group Mean ($\eta_{p,r,q}^t$):} The normalized group mean is defined as $\eta_{p,r,q}^t = 100 \times (\bar{x}_{p,r,q}^t - x^*_{q})/(x^*_{q})$, where $x^*_{q}$ is the ground truth value for question $q$. This metric provides a relative measure of the group's average estimate compared to the ground truth, normalized by the scale of each question. We further scale by 100 to express the group mean as the \textit{percentage} of ground truth $x^*$; $+100$ means overestimating by one unit of $x^*$, and $-100$ means underestimating by one unit of $x^*$. Note that, unlike \citet{becker2019wisdom}, who use log functions for normalization for some questions, we adopt their ``alternative normalization procedure'' due to occasional negative values from LLM agents, as outlined in their supplementary materials. They demonstrate that various normalization methods yield similar results.

\paragraph{Normalized Group Error ($\varepsilon_{p,r,q}^t$):} The normalized group error, $\varepsilon_{p,r,q}^t$, measures the absolute deviation of the normalized group mean from the ground truth value and is defined as $\varepsilon_{p,r,q}^t = |\eta_{p,r,q}^t| = 100 \times |(\bar{x}_{p,r,q}^t - x^*_{q})/x^*_{q}|$. Note that we also scale by 100, so $\varepsilon_{p,r,q}^t$ can be interpreted as the error in terms of the percentage of the ground truth value $x^*$. For example, $\varepsilon_{p,r,q}^t=50$ means that the group mean is deviating from the ground truth $x^*$ by $50\%$ of $|x^*|$.

\paragraph{Average Normalized Group Error ($\overline{\varepsilon_{t}}$):} To evaluate group performance across political leanings, runs, and questions, we calculate the \textit{average normalized group error}, denoted as $\overline{\varepsilon_{t}}$. This metric represents the average normalized group error at a specific time step $t$. Formally, $\overline{\varepsilon_{t}} = \frac{1}{P \cdot R \cdot Q} \sum_{p} \sum_{r=1}^R \sum_{q=1}^{Q} \varepsilon_{p,r,q}^t$,  where $P=2$ represents the number of political leanings, $R$ is the total number of runs, and $Q=8$ is the number of questions.

\paragraph{Group Error Reduction ($\Delta \varepsilon_{p,r,q}$):} The change in normalized group error from the initial to the final estimate is quantified as $\Delta \varepsilon_{p,r,q} = (\varepsilon_{p,r,q}^{t=3} - \varepsilon_{p,r,q}^{t=1})$, indicating error reduction for each run. Note that because $\varepsilon_{p,r,q}^t$ is already scaled by 100, $\Delta \varepsilon_{p,r,q}$ can be interpreted as the error reduction in terms of the percentage of the ground truth value $x^*$. For example, $\Delta \varepsilon_{p,r,q}=-50$ means that the group error $\delta_{p,r,q}^t$ reduces by $50\%$ of the size of the ground truth $|x^*|$. In other words, the group mean $x_{p,r,q}^t$ moves towards the ground truth $x^*$ by $50\%$ of the size of the ground truth $|x^*|$)

\paragraph{Average Group Error Reduction ($\overline{\Delta \varepsilon}$):} To quantify the change in group error over the course of the experiment, we calculate the \textit{average group error reduction}, denoted as $\overline{\Delta \varepsilon}$. This is calculated by averaging $\Delta \varepsilon_{p,r,q}$ across all political leanings, runs, and questions. Formally, $\overline{\Delta \varepsilon} = \frac{1}{P \cdot R \cdot Q} \sum_{p} \sum_{r=1}^{R} \sum_{q=1}^{Q} \Delta \varepsilon_{p,r,q}$, where $P=2$ represents the number of political leanings, $R=12$ the total number of runs, and $Q=8$ the number of questions. $\overline{\Delta \varepsilon}$ reflects the \textit{wisdom of partisan crowds} effect in LLM agents. A more negative value of $\overline{\Delta \varepsilon}$ indicates a stronger wisdom of partisan crowds effect.

\subsection{Deriving the Revision Coefficient}\label{app:revision_coefficient}
To compute the revision coefficient, we follow the methodology outlined by \citet{becker2017network}, focusing on the relationship between individual revisions, initial errors, and the social signal. The process involves two main steps: regression to obtain residuals and computing the partial correlation.

\paragraph{Obtaining Residuals:}
The first step involves using Ordinary Least Squares (OLS) regression to adjust individual revisions and errors based on the social signal. This is done to isolate the effect of an individual's initial accuracy on their subsequent revision, independent of the strength of the social influence they experience.

\begin{enumerate}
    \item \textbf{Adjusting Individual Revisions:} For each question $q$, across all individual $i$, political leaning $p$, and run $r$, we perform a regression of their individual revision $\Delta x_{i,p,r,q}$ against the social signal $s_{i,p,r,q}$. For a given question $q$, the regression equation is:
    \begin{align*}
        \Delta x_{i,p,r,q} = a_{1,q} \cdot s_{i,p,r,q} + b_{1,q} + \epsilon_{1,i,p,r,q} 
    \end{align*}
       where $a_{1,q}$ and $b_{1,q}$ are regression coefficients, and $\epsilon_{1,i,p,r,q}$ is the residual. The residual $\epsilon_{1,i,p,r,q}$ represents the adjusted individual revision $\widetilde{\Delta x}_{i,p,r,q}=\epsilon_{1,i,p,r,q}$.

    \item \textbf{Adjusting Initial Errors:} Similarly, we regress the individual initial error $e_{i,p,r,q}$ against the same social signal:
        \begin{align*}
            e_{i,p,r,q} = a_{2,q} \cdot s_{i,p,r,q} + b_{2,q} + \epsilon_{2,i,p,r,q} 
       \end{align*}
       where $a_{2,q}$ and $b_{2,q}$ are coefficients, and $\epsilon_{2,i,p,r,q}$ is the residual. The residual $\epsilon_{2,i,p,r,q}$ becomes the adjusted initial error $\widetilde{e}_{i,p,r,q}=\epsilon_{2,i,p,r,q}$.
\end{enumerate}

\paragraph{Computing the Partial Correlation:}
With the residuals obtained, the revision coefficient ($r_{\text{adj},q}$) is computed as the Pearson correlation between the adjusted individual revisions and adjusted initial errors:

   \begin{align*}
    r_{\text{adj},q} &= \text{corr}(\Delta \widetilde{x}_{i,p,r,q},\widetilde{e}_{i,p,r,q}) \\
    &= \text{corr}(\epsilon_{1,i,p,r,q}, \epsilon_{2,i,p,r,q})
\end{align*}

This partial correlation reflects the extent to which individuals with higher initial accuracy are less influenced by social signals in revising their estimates, after controlling for the strength of the social signal itself.

\end{document}